\documentclass[letterpaper, 10 pt, journal, twoside]{IEEEtran}
\usepackage{xcolor,soul,framed} %,caption

\colorlet{shadecolor}{yellow}
\usepackage[pdftex]{graphicx}
\graphicspath{{../pdf/}{../jpeg/}}
\DeclareGraphicsExtensions{.pdf,.jpeg,.png}
\usepackage{microtype} 
\usepackage{booktabs,multirow}
\usepackage{makecell}
\usepackage{amsmath}
\usepackage{array}
\usepackage{float}
\usepackage{color}
\usepackage{multirow}
\usepackage{stfloats}
\usepackage{subfigure,bbm}
\usepackage{amsfonts,amssymb}
\usepackage[colorlinks,linkcolor=blue,anchorcolor=blue,citecolor=blue]{hyperref}
\begin{document}
%\bstctlcite{IEEE:BSTcontrol}
\IEEEoverridecommandlockouts
\overrideIEEEmargins

%=== TITLE & AUTHORS ====================================================================Learning Task-adaptive Quasi-stiffness Control\\ for a Powered Transfemoral Prosthesis
\title{\LARGE \bf
A Learning Quasi-stiffness Control Framework of a Powered Transfemoral Prosthesis for Adaptive Speed and Incline Walking}
% \bstctlcite{IEEEexample:BSTcontrol}
  \author{Teng Ma$^*$,
        Shucong Yin$^*$,
        Zhimin Hou$^*$,
        Yuxuan Wang,
        Binxin Huang,
        Haoyong Yu,
        and Chenglong Fu%   <-this % stops a space
  \thanks{This work was supported by the National Natural Science Foundation of China [Grant U1913205]; the Science, Technology and Innovation Commission of Shenzhen Municipality [Grant ZDSYS20200811143601004]; the Stable Support Plan Program of Shenzhen Natural Science Fund [Grant 20200925174640002]; the Science and Engineering Research Council, Agency of Science, Technology and Research, Singapore, through the National Robotics Program under Grant No.M22NBK0108; and Centers for Mechanical Engineering Research and Education at MIT and SUSTech. (Corresponding author: Chenglong Fu) (Teng Ma, Shucong Yin, and Zhimin Hou contributed equally to this work.)}
  \thanks{Shucong Yin, Yuxuan Wang, Binxin Huang, and Chenglong Fu are with Shenzhen Key Laboratory of Biomimetic Robotics and Intelligent Systems and Guangdong Provincial Key Laboratory of Human Augmentation and Rehabilitation Robotics in Universities, Department of Mechanical and Energy Engineering, Southern University of Science and Technology, Shenzhen, 518055, China. (e-mail: fucl@sustech.edu.cn;bieyhy@nus.edu.sg)}% <-this % stops a space
  \thanks{Teng Ma, Zhimin Hou, and Haoyong Yu are with the Department of Biomedical Engineering, National University of Singapore, Singapore 119077, Singapore.}
}

\markboth{IEEE TRANSACTIONS ON COGNITIVE AND DEVELOPMENTAL SYSTEMS}%
{Ma \MakeLowercase{\textit{et al.}}: Tuning-free Quasi-stiffness Control framework of a Powered Transfemoral Prosthesis for Task-adaptive Walking}
% \markboth{ICRA}{Ma \MakeLowercase{\textit{et al.}}: Learning Task-adaptive Quasi-stiffness for a Powered Transfemoral Prosthesis Control}

% ====================================================================
\maketitle
% \thispagestyle{empty}
% \pagestyle{empty}
% === ABSTRACT ====================================================================
% 
\begin{abstract}
Impedance-based control represents a prevalent strategy in the powered transfemoral prostheses because of its ability to reproduce natural walking. However, most existing studies have developed impedance-based prosthesis controllers for specific tasks, while creating a task-adaptive controller for variable-task walking continues to be a significant challenge. This article proposes a task-adaptive quasi-stiffness control framework for powered prostheses that generalizes across various walking tasks, including the torque-angle relationship reconstruction part and the quasi-stiffness controller design part. 
A Gaussian Process Regression (GPR) model is introduced to predict the target features of the human joint's angle and torque in a new task. Subsequently, a Kernelized Movement Primitives (KMP) is employed to reconstruct the torque-angle relationship of the new task from multiple human reference trajectories and estimated target features. Based on the torque-angle relationship of the new task, a quasi-stiffness control approach is designed for a powered prosthesis. Finally, the proposed framework is validated through practical examples, including varying speeds and inclines walking tasks. 
Notably, the proposed framework not only aligns with but frequently surpasses the performance of a benchmark finite state machine impedance controller (FSMIC) without necessitating manual impedance tuning and has the potential to expand to variable walking tasks in daily life for the transfemoral amputees.

% The proposed framework meets and often surpasses the performance of a benchmark finite state machine impedance controller (FSMIC) without necessitating manual impedance tuning and has the potential to expand to variable walking tasks in daily life for the transfemoral amputees.
\end{abstract}

% === KEYWORDS ==================================================

% \begin{IEEEkeywords}
% \hl{}
% \end{IEEEkeywords}

% \fi
% \IEEEpeerreviewmaketitle

% === I. INTRODUCTION =============================================================

\section{Introduction}
% % =======
% % FIG. 01sarkisian2020design
% \IEEEPARstart{T}{ransfemoral} amputees suffer ambulation difficulties when performing demanding locomotion modes, such as climbing ramps and stairs. Powered transfemoral prostheses have the potential to restore the lost function of the joints and enhance the mobility of the amputees\cite{gehlhar2023review}. Significant progress has been made in both the mechanical design \cite{sup2008design,lawson2014robotic,elery2020design,tran2022lightweight} and controller design\cite{gregg2014virtual,li2021toward, mendez2020powered} of powered prostheses over the past decades. Transfemoral amputees benefit from these powered prostheses that may improve gait symmetries, increase walking speed, and reduce metabolic expenditure\cite{gehlhar2023review}. 
% Impedance-based control strategies are the most commonly used methods because they can simply reproduce natural walking. These methods segment the gait cycle into several sub-phases with each sub-phase performing an impedance control law. 
% Researchers manually tuned impedance parameters to achieve task-specific tasks, such as flat walking\cite{sup2008design}, ramp climbing\cite{sup2010upslope}, and stair climbing\cite{lawson2012control}. However, the heuristic parameter tuning procedure for a specific task takes at least several hours. 
Individuals with transfemoral amputations suffer difficulties in ambulation, particularly during demanding locomotion activities such as ascending ramps and stairs. Powered transfemoral prostheses hold promise in restoring the lost joint function and significantly enhancing the mobility of these individuals \cite{hong2023feasibility,gehlhar2023review}. While significant progress has been made in both the mechanical designs \cite{sup2008design,feng2020energy,elery2020design,azocar2020design,tran2022lightweight,zhu2022design} and controller designs \cite{gregg2014virtual,pi2020biologically,kim2022deep,li2021toward,mendez2020powered,wang2022design,gehlhar2022powered,nuesslein2024deep} of powered prostheses over the past years, developing a prosthesis control framework that generalizes across various locomotion modes and terrain conditions remains a challenge \cite{gehlhar2023review}. 

Impedance-based control is the most commonly used approach in powered prostheses because of its ability to reproduce natural walking. These methods segment the gait cycle into several sub-phases, each performing an impedance control law. Previous researchers manually tuned the impedance parameters for each task, such as flat walking \cite{sup2008design}, ramp climbing \cite{sup2010upslope}, and stair climbing \cite{lawson2012control}. However, the heuristic parameter tuning procedure for a specific task takes at least several hours \cite{best2023data}. Transfemoral amputees may perform different tasks (locomotion modes and terrain conditions) in daily walking. The time required to tune a powered prosthesis for each task and each patient presents a significant barrier to its clinical feasibility \cite{lenzi2014speed}. To reduce the tuning time, reinforcement learning (RL) is implemented in \cite{li2021toward,wen2019online} to tune the impedance parameters automatically. Nevertheless, these approaches are limited to flat walking task and knee joint only at present. 
% A data-driven controller that uses variable impedance control during stance is present in \cite{best2023data} for varying speeds and inclines walking. () Reinforcement learning is implemented in \cite{li2021toward} to automatically tune the impedance parameters. Although it can reduce the manual tuning time, these approaches are limited to the knee joint only and flat walking tasks at present. 
Recent studies have achieved prostheses handling variable-task walking. Data-driven, impedance-based controllers are respectively designed for varying speeds and inclines walking \cite{best2023data}, stair climbing \cite{cortino2023data}, and sitting/standing \cite{welker2022data}. An adaptive stair controller enables transfemoral amputee patients to climb stairs of varying heights \cite{hood2022powered}. These controllers are still required to extend to a unified framework for variable-task walking. To date, developing powered transfemoral prostheses controllers that generalize across tasks remains an open problem. 
% \begin{figure}[!t]
%   \begin{center}
%   \includegraphics[width=0.8\linewidth]{pdf/fig1.pdf}\\
%   \caption{A variety of tasks of transfemoral amputees walking with powered transfemoral prostheses, such as different locomotion modes (flat walking, ramp climbing, and stair climbing) with different terrain conditions (speeds, ramp inclinations, and stair heights). Designing controllers for powered prostheses that generalize across tasks is challenging and necessary.}
%   \label{fig:1}
%   \end{center}
%     \vspace{-0.6cm}
% \end{figure}
%these controllers are still required to extend to a unified framework for the variety of tasks that transfemoral amputee patients may perform in daily walking.

%The time needed to tune a powered transfemoral prosthesis for each patient has been identified as a main obstacle to their clinical viability. 
% % =======
\begin{figure*}[!t]
  \begin{center}
  \includegraphics[width=1.0\linewidth]{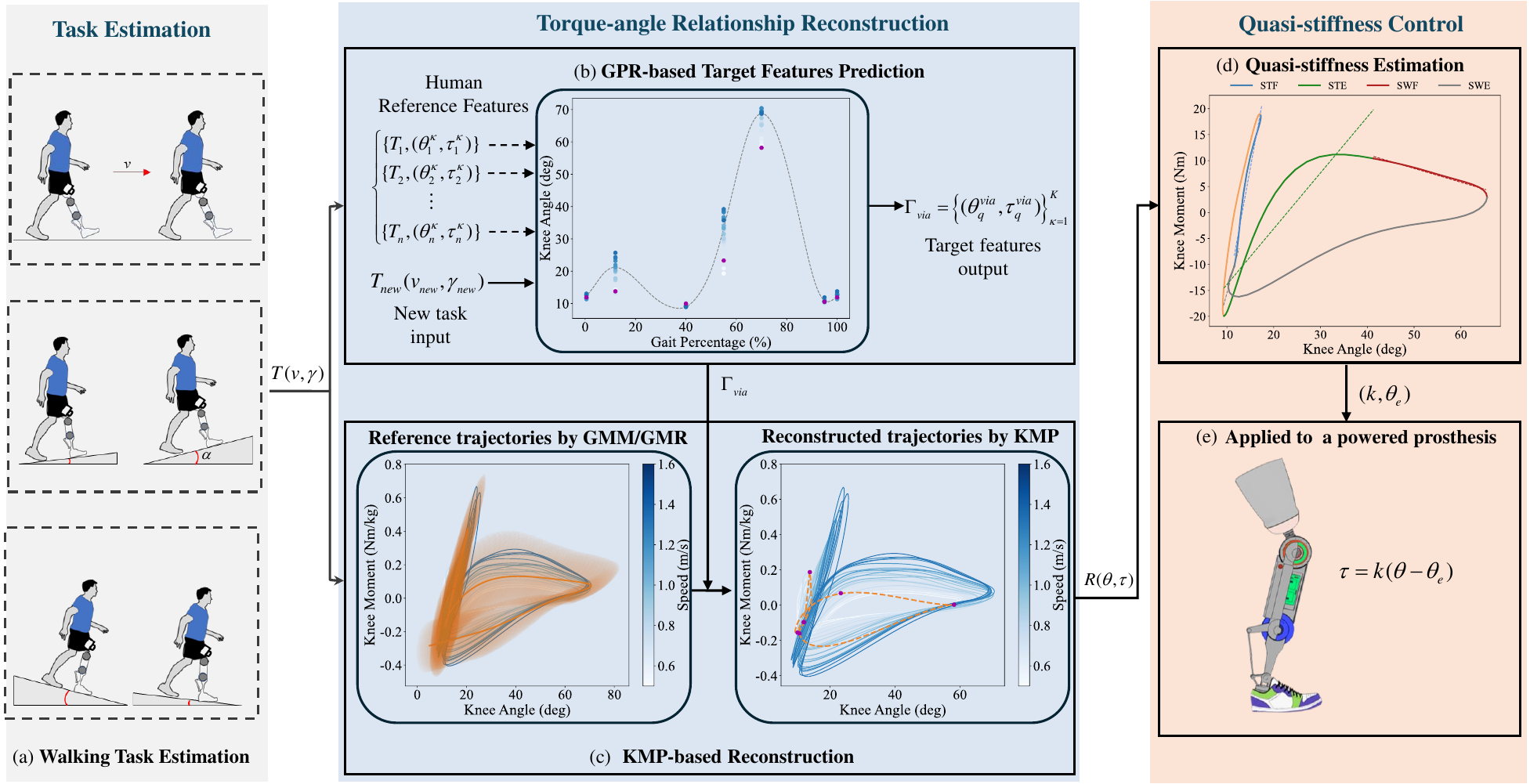}
  \caption{Overview of the proposed learning quasi-stiffness control framework. (a) Task estimation as input for learning the new torque-angle relationship. $T(v,\gamma)$ denotes the task parameter (where $v$ denotes the walking speed, $\gamma$ represents the terrain parameters). (b) Diagram of GPR-based target features prediction. Typically, the lower limb joints' angle and torque trajectories in one gait cycle have local maximum and minimum, which can be selected as target features to represent the kinematic and kinetic of the joint in one specific task \cite{li2021toward}. The blue dots ($\theta_i^\kappa$, where $\theta_i^\kappa$ denotes the $\kappa$ features of the $i$-th reference) represent the target features from human reference trajectories, and the purple dots indicate the predicted target features (${\Gamma _{via}} = \{ \theta _q^{via},\tau _q^{via}\} _{\kappa = 1}^K,$ where $K$ represents the number of target features) of a new task. (c) KMP-based torque-angle relationship reconstruction. The left figure represents the reference torque-relationship retrieved by GMM/GMR and the right figure denotes the reconstructed torque-angle relationship $R(\theta, \tau)$ by the KMP. (d) Quasi-stiffness estimation from the reconstructed torque-angle relationship. Each dashed line represents a linear regression of each part of the relationship. The slope of the linear regression function denotes the quasi-stiffness. The intersection point of the regression equation with the x-axis represents the equilibrium angle. (e) Applying the quasi-stiffness control on a powered transfemoral prosthesis.}
  \label{fig:1}
  \end{center}
\end{figure*}
Quasi-stiffness control is a subset of the impedance-based control strategies \cite{gehlhar2023review}. While traditional impedance control requires tuning numerous impedance parameters to generate biologically joint torques, this quasi-stiffness strategy determines the impedance parameters of each sub-phase from the torque-angle relationship of human lower limb joints. The slope of the torque-angle relationship is called "quasi-stiffness" \cite{gehlhar2023review,rouse2012difference}. The desired prosthesis torque is then determined by the quasi-stiffness $k$ and equilibrium angle $\theta_e$ as $\tau (\theta ) = k(\theta  - {\theta _e})$. The torque–angle relationships can be modified for quasi-stiffness control of powered prostheses in different tasks. For example, the stair controller modulates the torque-angle relationship based on the stair height for stair climbing tasks \cite{hood2022powered}. While quasi-stiffness control strategies have been proven can improve gait symmetry and achieve speed-adaption without speed-specific tuning \cite{lenzi2014speed}, how to adapt to other tasks automatically is still a challenge.
% In addition, it is reported that quasi-stiffness control strategies have the potential to improve gait symmetry and achieve speed-adaption without subject-specific or speed-specific tuning \cite{lenzi2014speed}. 

Inspired by the previous research on quasi-stiffness control strategies, this study proposes a tuning-free quasi-stiffness control framework that generalizes across a variety of tasks. Firstly,  a novel task-adaptive torque-angle relationship reconstruction method is proposed for autonomously generating the torque-angle relationship of a new task. The torque-angle relationship is constructed by the the joint angle and torque and its profile depends strongly on the peak values of the joint angle and torque. The peak values (named as target features in this study) in the profiles of joint angles and torques vary with tasks and are often used to represent the joint kinematics and kinetics during walking tasks \cite{li2021toward,wen2019online}. However, most torque-angle relationships and target features are obtained in the laboratory by a motion capture system. It is only possible to collect joint kinematics and kinetics for some tasks. This study proposes an imitation learning-based method that can reconstruct the torque-angle relationships of new tasks by learning from multiple reference torque-angle relationships. For example, reconstructing a relationship at a new walking speed from learning the existing reference relationships at different speeds (or a new relationship at a new ramp inclination, etc.). To reconstruct the torque-angle relationship of a new task, the Gaussian Process Regression (GPR) \cite{williams2006gaussian,schulz2018tutorial} is introduced to predict the target features of the joint angles and torques based on the reference features and task variables (such as speeds and terrain parameters). Considering the limited human reference size, GPR, a nonparametric kernel-based probabilistic model, adeptly balances fitting accuracy with model complexity, mitigating the risk of over-fitting in small-size training samples.
Furthermore, the torque-angle relationship of the new task can be reconstructed by the target features and human reference torque-angle relationships. While many fitting methods can use these target features for torque-angle relationship reconstruction, such as polynomial interpolation \cite{koopman2014speed} and Fourier Series fitting \cite{embry2018modeling}, these methods consider only trajectory fitting instead of the multiple reference relationships. Instead, learning from multiple reference trajectory methods, such as probabilistic trajectory methods can be introduced. Gaussian Mixture Model/Regression (GMM/GMR) \cite{cohn1996active,calinon2016tutorial} was applied to reproduce joint angles and learning trajectory distribution through learning from the reference database. However, this method makes it difficult to modify the learned trajectories to fulfill additional constraints, such as passing through specific target features. A Kernelized Movement Primitives (KMP) \cite{ huang2019kernelized} method was presented that allows the reproduced learning from reference trajectories passing through the specific via-points. Inspired by this, KMP is employed to construct the new torque-angle relationship from multiple reference relationships and target features. Secondly, the quasi-stuffiness control strategy is introduced after obtaining the torque-angle relationship to design a task-adaptive controller that generalizes across tasks.

This study aims to develop a task-adaptive, tuning free quasi-stiffness control framework for a powered transfemoral prosthesis that can reproduce natural walking and biomimic joint kinematics and kinetics similar to those of able-bodied people. The main contributions are summarized as follows:
\begin{enumerate}
\item A learning-based \emph{task-adaptive} torque-angle relationship construction method is proposed for powered transfemoral prostheses, which can autonomously generate each task-specific torque-angle relationship during continuously varying tasks. A \emph{tuning-free} quasi-stiffness control method based on the learned torque-angle relationship is presented for powered transfemoral prostheses.
\item Experimental results of varying speed and varying
incline walking tasks show that the proposed control framework reproduces biomimetic kinematics and kinetics similar to able-bodied references, meets the performance of the most widely used (one of the state
of the art) FSMIC without requiring manual impedance tuning, and has the potential to enable autonomous task adaptation during continuously varying walking tasks.
\end{enumerate}

\section{Method}
% As shown in Fig. \ref{fig:2}, the proposed task-adaptive learning quasi-stiffness control framework consists of a learning from human demonstrations part and a quasi-stiffness control part. The learning-based part is composed of two models: a GPR-based target features prediction model and a KMP-based task-specific torque-angle relationship reconstruction model.
%考虑要不要加
\subsection{Overview of the Framework}
As shown in Fig.~\ref{fig:1}, the proposed task-adaptive quasi-stiffness control framework consists of a task estimation part, a torque-angle relationship reconstruction part, and a quasi-stiffness control part. The task estimation part perceives amputee locomotion information and terrain
conditions. It provides task parameters $T(v,\gamma)$ ( where $v$  denotes walking speed and $\gamma$ represents terrain parameters, such as inclination $\alpha$ and stair height $h$.) for the next part. Previous research, including our own, has explored a variety of task estimation methodologies, encompassing vision-based approaches \cite{zhang2019environmental,zhong2020environmental} for assessing terrain conditions and IMU (Inertial Measurement Unit)-based strategies \cite{best2023data,young2013training,ma2022piecewise} for estimating walking speed. These methodologies facilitate the estimation of task parameters $T(v,\gamma)$. In this study, we used the speed estimation method in ~\cite{best2023data,ma2022piecewise} and the incline estimation method in~\cite{zhang2019environmental,zhang2020subvision}. The task estimate accuracy is shown in Table~\ref{tab:1}.
\begin{table}[h]
    \centering
    \caption{Task estimate accuracy during the varying task trials}
    {
\begin{tabular}{cccccc}
   \hline\noalign{\smallskip}
   Task & Accuracy & Reference Method\\
  \hline\noalign{\smallskip}
    Speed& 94 \% & \cite{best2023data,ma2022piecewise} \\
    Incline& 95 \% & \cite{zhang2019environmental,zhang2020subvision}\\
 \hline
\end{tabular}}
    \label{tab:1}
\end{table} 

The torque-angle relationship reconstruction part includes a GPR-based target features prediction model and a KMP-based task-specific torque-angle relationship reconstruction model. This part automatically constructs the joints' torque-angle relationship in a new task. The quasi-stiffness control part determines the parameters and applies to a quasi-stiffness controller based on the generated torque-angle relationship. The details of these two parts are as follows.

% Previous research, including our own, has explored a variety of task estimation methodologies, encompassing vision-based approaches \cite{zhang2019environmental,zhong2020environmental} for assessing terrain conditions and IMU (Inertial Measurement Unit)-based strategies \cite{best2023data,young2013training,ma2022piecewise} for estimating walking speed. These methodologies facilitate the estimation of task parameters $T(v,\gamma)$, where $v$  denotes walking speed and $\gamma$ represents terrain parameters, such as inclination $\alpha$ and stair height $h$.

% Our previous studies and some other studies presented various task estimation methods, such as vision-based methods \cite{zhang2019environmental,zhong2020environmental} for terrain condition estimation and IMU (inertial measurement unit)-based methods \cite{best2023data,young2013training,ma2022piecewise} for speed estimation. These methods can be applied to estimate the task parameter
% $T(v,\gamma)$ (where $v$ denotes the walking speed, $\gamma$ represents the terrain parameters, such as the inclinations $\alpha$ or the stair heights $h$) for the proposed framework.

\subsection{Target Features Prediction} 
% % =======
% % FIG. 03
% % =======
% \begin{figure}[!t]
%   \begin{center}
%   \includegraphics[width=1.0\linewidth]{pdf/fig3.pdf}\\
%   \caption{Diagram of GPR-based target features prediction. Typically, the lower limb joints' angle and torque trajectories in one gait cycle have local maximum and minimum, which can be selected as target features to represent the kinematic and kinetic of the joint in one specific task \cite{li2021toward}. The blue dots represent the target features from human demonstrations, and the purple dots indicate the predicted target features of a new task.}
%   \label{fig:3}
%   \end{center}
%   \vspace{-0.5cm}
% \end{figure}

% the Gaussian Process Regression (GPR) \cite{williams2006gaussian,schulz2018tutorial} is introduced to predict the target features of the joint angles and torques based on the demonstrated features and task variables (such as speed and terrain parameters).

\begin{figure}[!t]
  \begin{center}
  \includegraphics[width=0.8\linewidth]{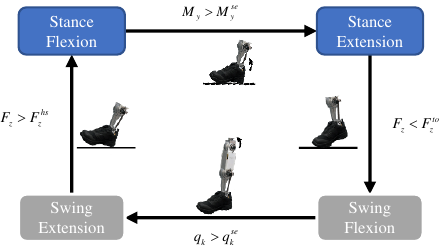}\\
  \caption{Finite state machine (FSM) of the quasi-stiffness control strategy. Similar to standard impedance control, the quasi-stiffness control strategy also divides the gait cycle into four phases. Stance extension begins when the measured moment $M_y$ reaches the threshold $M_y^{se}$. Once the measured vertical ground reaction force $F_z$ is less than a threshold $F_z^{to}$, which means toe-off happens, the stance extension state switches to the swing flexion state. When the knee angle $q_k$ reaches the swing extension threshold $q_k^{se}$, the swing extension state starts. Finally, the FSM start backs at stance flexion state when the $F_z$ is greater than $F_z^{hs}$ (heel strike happens).}
  \label{fig:2}
  \end{center}
\end{figure}

Two open-source dataset \cite{camargo2021comprehensive,reznick2021lower} are implemented as the human reference trajectories, which contains tasks including multiple locomotion modes (flat walking and ramp climbing) and terrain conditions (walking speeds (28 speeds ranging from 0.5 to 1.85 m/s in 0.05 m/s increments) and ramp inclinations (6 inclination angles in \cite{camargo2021comprehensive} and 4 inclination angle in \cite{reznick2021lower}), rang from -10 deg to 10 deg with 5 deg increments). Notably, for varying walking speeds on treadmill, we divided 70 \% of the data as training set and 30 \% as testing set; for varying inclines, we used all the incline walking data in \cite{camargo2021comprehensive} and \cite{reznick2021lower} as training set and the data we collected under the motion capture system (rang from -8 deg to 8 deg in 2 deg increments) as testing set. To obtain the new task-specific torque-angle relationship, this study introduces GPR-based learning from the reference trajectory method for target features prediction, as shown in Fig. \ref{fig:2}(b).
% Considering the limited human demonstration size, GPR, a nonparametric kernel-based probabilistic model, adeptly balances fitting accuracy with model complexity, mitigating the risk of over-fitting in small-size training samples. 
Assume there are $n$ reference trajectories and define the train data as
\begin{equation}
\begin{array}{cc}
\boldsymbol{x}_i=T_i=(v_i, \gamma_i),\\
\boldsymbol{D}_\kappa=(\boldsymbol{x}_i, \boldsymbol{y}(\kappa, i)), \kappa=1,2,\cdots,6, i= 1,2,\cdots,n,
\end{array}
\end{equation}
where $\kappa$ denotes the number of the features; $\boldsymbol{x}_i$ is the task parameter of the $i$-th reference trajectory ($i$-th task). $\boldsymbol{y}(\kappa, i)$ is the output features (joint angles/torques) of the $i$-th reference trajectory. For the task $\boldsymbol{x}_i$, the GPR \cite{williams2006gaussian,schulz2018tutorial} is intended to transform the input task vector $T_i$ into a output value $\boldsymbol{y}(\kappa, i)$ by $\boldsymbol{y}(\kappa, i)=f_i(\boldsymbol{x}_i)+\epsilon_i$, where $\epsilon_i$ represents the Gaussian noise satisfying ${\epsilon _i} \sim {\cal N}(0,{\sigma _\epsilon^2})$.  
The training dataset can be described by the GPR model as
\begin{equation}
{f_\kappa }({\boldsymbol{x}}) \sim {\cal G}{\cal P}(\mu_\kappa({\boldsymbol{x}}),\phi_\kappa({\boldsymbol{x}},{\boldsymbol{x}'})),
\end{equation}
where $\mu_\kappa(\boldsymbol{x})$ represents the mean function and is set to zero \cite{schulz2018tutorial};
$\phi_\kappa({\boldsymbol{x}},{\boldsymbol{x}'})$ denotes the covariance function (also called Gaussian kernel) and can be selected in \cite{williams2006gaussian}. 
% is frequently taken 
% \begin{equation}
% k(\boldsymbol{x},\boldsymbol{x}') = {\sigma _s^2}\exp \left( { - \frac{1}{2}{{(\boldsymbol{x} - \boldsymbol{x}')}^T}\lambda (\boldsymbol{x} - \boldsymbol{x}')} \right),
% \end{equation}
% where $\sigma_s^2$ is the signal variance and $\lambda$ the width-scale of the Gaussian kernel. 
The target features' prior distribution in $\boldsymbol{D}_\kappa$ satisfies
\begin{equation}
\boldsymbol{y} \sim {\cal N}(\boldsymbol{0},\mathbf{\Phi(X,X)}+\sigma_\epsilon^2 \boldsymbol{I}),
\end{equation}
where $\mathbf{\Phi(X,X)}$ is the covariance matrix. To generate prediction for a new task input $\boldsymbol{x}_*$, the predicted function $f_*(\boldsymbol{x}_*)$ is following a joint Gaussian distribution:
\begin{equation}
\left[\! {\begin{array}{*{20}{c}}
\boldsymbol{y}\\
{{f_*}(\boldsymbol{x}_*)}
\end{array}} \!\right] \!\sim \! {\cal N}\left(\! {\boldsymbol{0},\left[ {\begin{array}{*{20}{c}}
{\mathbf{\Phi(X,X)} + {\sigma _\varepsilon }\boldsymbol{I}}&{\mathbf{\Phi}(\mathbf{X},\boldsymbol{x}_*)}\\
{\mathbf{\Phi}(\boldsymbol{x}_*,\mathbf{X})}&{\phi({\boldsymbol{x}_*},{\boldsymbol{x}_*})}
\end{array}} \right]} \!\right)\!.
\end{equation}
The posterior distribution of the prediction$f_*(\boldsymbol{x}_*)$ at a new input $\boldsymbol{x}_*$ is  given by
\begin{equation}
p({f_*}|\boldsymbol{D}_\kappa,\boldsymbol{x}_*) \sim {\cal N}({\mu_*},\sigma_{{f_*}}^2).  
\end{equation}
The predicted mean value ${\mu_*}$ and corresponding variance ${\sigma_{f_*}} $ of the conditional distribution can be given by \cite{schulz2018tutorial} 
\begin{equation}
\begin{array}{l}
{u_*}(\boldsymbol{x}_*) = \mathbf{\Phi}_*{(\mathbf{\Phi} + \sigma _\varepsilon ^2\boldsymbol{I})^{ - 1}}\boldsymbol{y} = \mathbf{\Phi}_*\boldsymbol{\alpha},\\
\sigma_{\!{f_ *}\!}^2(\boldsymbol{x}_*) = \phi(\boldsymbol{x}_ *,\boldsymbol{x}_*) - {\mathbf{\Phi}_*}{(\mathbf{\Phi} + \sigma_\varepsilon ^2\boldsymbol{I})^{-1}}\mathbf{\Phi}_*^T,
\end{array}
\end{equation}
where $\mathbf{\Phi}=\mathbf{\Phi(X, X)}$, $\mathbf{\Phi}_*=\mathbf{\Phi}(\mathbf{X},\boldsymbol{x}_*)$ denotes the covariance matrix between training and new input, and $\phi({\boldsymbol{x}_*},{\boldsymbol{x}_*})$ denotes the covariance between the new input vector. 

Thus, the new output $f_*(\boldsymbol{x}_*)$ is predicted as the mean values. By employing the GPR method, the new task's target features including joint angles and joint torques can be represented as ${\Gamma _{via}} = \{ \theta _q^{via},\tau _q^{via}\} _{\kappa = 1}^K,$ where $K$ represents the number of target features.

\subsection{Torque-angle Relationship Reconstruction}
After obtaining the kinematics and kinetics target features of the new task, the KMP is employed to reconstruct the torque-angle relationship with the target features of the new task, as shown in Fig. \ref{fig:1}(c). Denote the set of reference torque-angle relationship by $\{ \{ {{\theta} _{m,n}},{{\tau}_{m,n}}\} _{m = 1}^M\} _{n = 1}^N,$ where ${\theta} _{m,n}\in \mathbb{R}^{1}$ represents the joint kinematics and ${\tau}_{m,n}\in \mathbb{R}^{1}$ is the joint kinetics. $M$ and $N$ denote the length of the torque-angle relationship and the number of the reference trajectories. GMM \cite{calinon2016tutorial} is introduced to encode the reference dataset and estimate the relationship probability distribution ${\cal P}(\theta, \tau)$
\begin{equation}
{\cal P}(\theta ,\tau ) = \left[ {\begin{array}{*{20}{c}}
\theta &\tau 
\end{array}} \right]^T \sim \sum\nolimits_{l = 1}^L {{\pi_l}} {\cal N}({\mu_l},{\Sigma_l}),
\end{equation}
where $\pi_l$, $\mu_l$, and $\Sigma_l$ represent prior probability, mean, and covariance, respectively. $L$ is the number of the Gaussian component. Furthermore, a probabilistic reference relationship $\{ {{\hat \theta }_m},{{\hat \tau }_m}\} _{m = 1}^M$ can be retrieved by GMR 
\begin{equation}
{{\cal P}_p}({{\hat \tau }_m}|{{\hat \theta }_m}) \sim {\cal N}({{\hat \mu }_m},{{\hat \Sigma }_m}),
\end{equation}
where $\hat \tau$ and $\hat \theta$ represent the reference torque and angle respectively, ${\hat \mu }_m$ and ${\hat \Sigma }_m$ denote mean and variance respectively. As mentioned before, we expect that the new torque-angle relationship can pass through the target features of the new task. Therefore, KMP is employed to learn from reference trajectores and fulfill the additional constraints. Considering a parametric torque-angle relationship can be formulated by KMP as
\begin{equation}
\tau (\theta ) = \Omega (\theta )w,
\end{equation}
where $\Omega(\theta)$ is a $\cal B$-dimensional basis function and is defined in \cite{huang2019kernelized}. $w$ is the weight vector and can be assumed to be normally distributed following $w \sim {\cal N}({\mu _w},{\Sigma _w})$. Thus, the probability distribution of the parametric torque-angle relationship stratifies
\begin{equation}
{{\cal P}_p}(\tau |\theta ) = {\cal N}(\Omega {(\theta )^T}{\mu _w},\Omega {(\theta )^T}{\Sigma _w}\Omega (\theta )).    
\end{equation}
The new torque-angle relationship is expected to pass through the target features. The target features is assumed satisfying $\tau _{q}^{via}|\theta _{q}^{via} \sim {\cal N}({\mu _v},{\Sigma _v})$ and 
\begin{equation}
{{\cal P}_v}(\tau_v|\theta_v ) = {\cal N}(\Omega {(\theta_v)^T}{\mu _v},\Omega {(\theta_v )^T}{\Sigma _v}\Omega (\theta_v )). 
\end{equation}
Finally, KMP requires matching the reference torque-angle relationship with the parametric torque-angle relationship and the target features. Inspired by \cite{huang2019kernelized,huang2020toward,zou2020learning}, to obtain $\mu_w$ and $\Sigma_w$, the \emph{Kullback-Leibler} (KL) divergence is introduced to design the cost function as 
\begin{equation}
\begin{array}{l}
J({\mu _w},{\Sigma _w}) = \sum\nolimits_{m = 1}^M {{D_{KL}}({{\cal P}_r}(\tau |{\theta _m})||{{\cal P}_p}(\tau |{\theta _m}))} \\
 \quad\quad\quad\quad\quad\quad+ \sum\nolimits_{q = 1}^Q {{D_{KL}}({{\cal P}_p}(\tau |{\theta _q})||{{\cal P}_q}(\tau |{\theta _q}))},
\end{array}
\end{equation}
where $D_{KL}$ denotes the KL divergence between the two probability distributions. A detailed solution to minimization of the cost function can be found in \cite{huang2019kernelized}. By employing the KMP, the torque-angle relationship of the new task can be reconstructed and applied in the quasi-stiffness control.

\begin{figure}[!t]
  \begin{center}
  \includegraphics[width=0.7\linewidth]{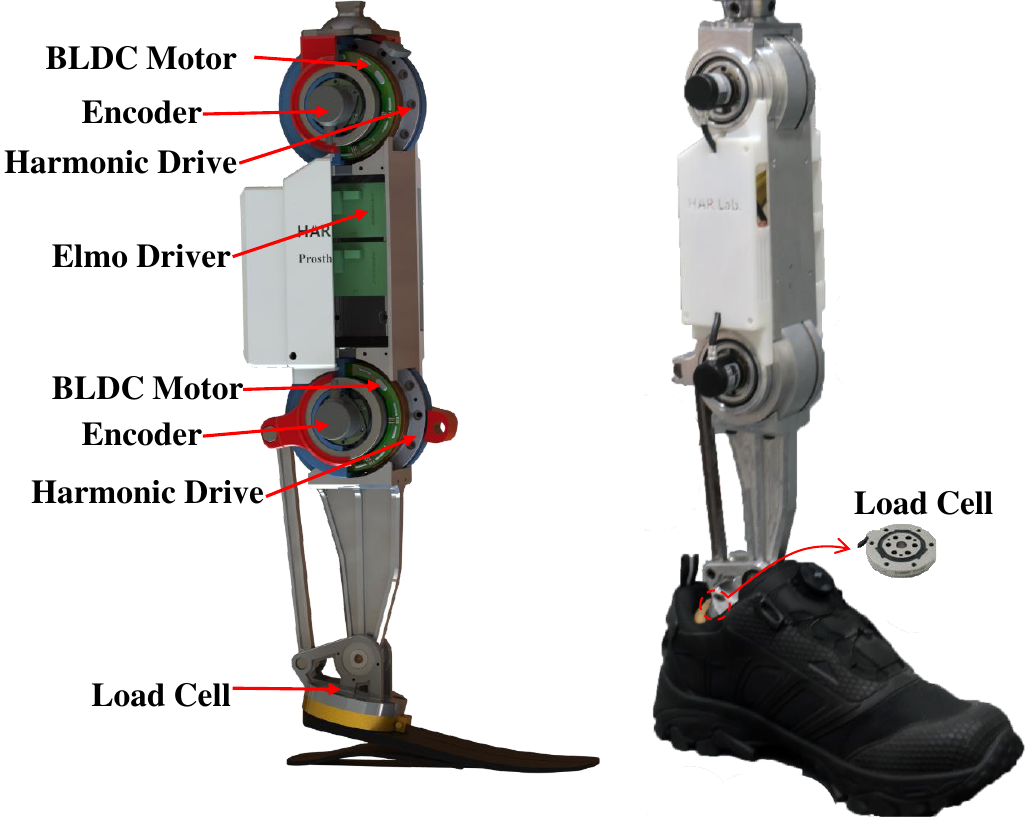}\\
  \caption{Powered transfemoral prosthesis used in this study: CAD version with
key components (left) and the real manufactured version (right).}
  \label{fig:3}
  \end{center}
\end{figure}

\subsection{Quasi-stiffness Estimation and Controller Design}
% \begin{figure}[!t]
%   \begin{center}
%   \includegraphics[width=1.0\linewidth]{pdf/fig4.pdf}\\
%   \caption{Quasi-stiffness estimated from the reconstructed torque-angle relationship. Each dashed line represents a linear regression of each part of the relationship. The slope of the linear regression function denotes the quasi-stiffness. The intersection point of the regression equation with the x-axis represents the equilibrium angle.}
%   \label{fig:4}
%   \end{center}
%     \vspace{-0.5cm}
% \end{figure}
\begin{figure}[!t]
  \begin{center}
  \includegraphics[width=1.0\linewidth]{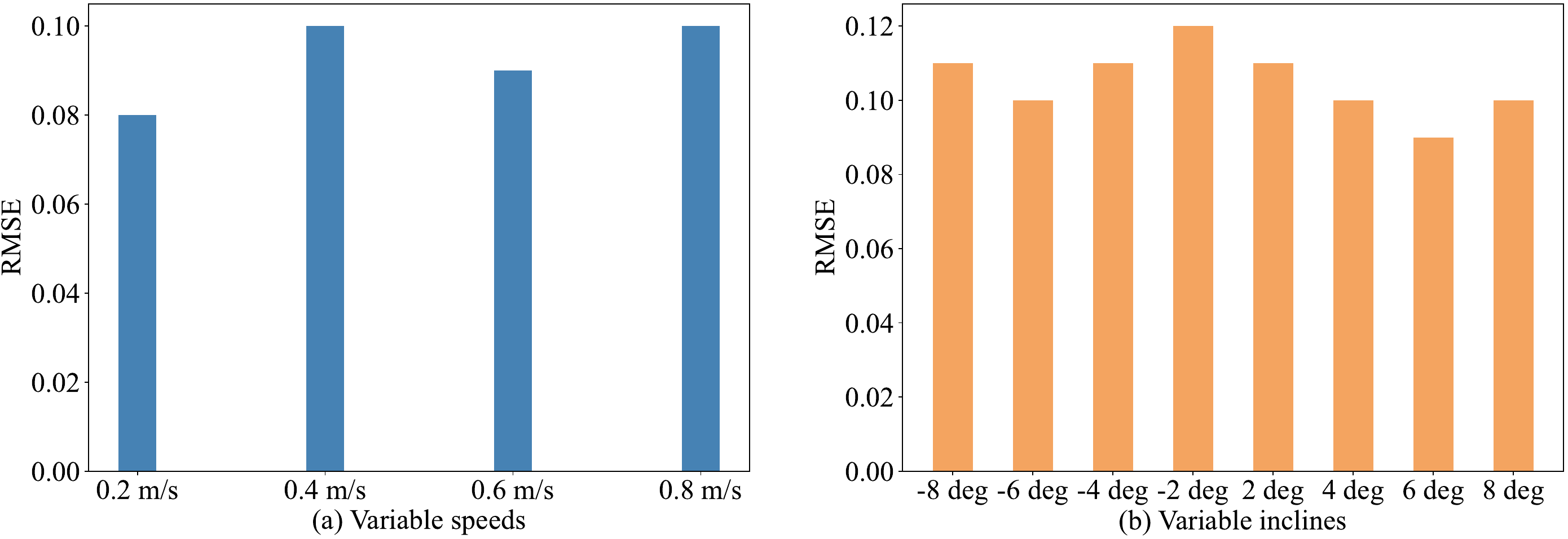}\\
  \caption{RMSE of the GPR model in the target feature estimation.}
  \label{fig:4}
  \end{center}
\end{figure}

\begin{figure*}[!t]
	\centering
	\subfigure[The reference torque-angle relationship retrieved by GMM/GMR.]{\includegraphics[width=0.49\linewidth]{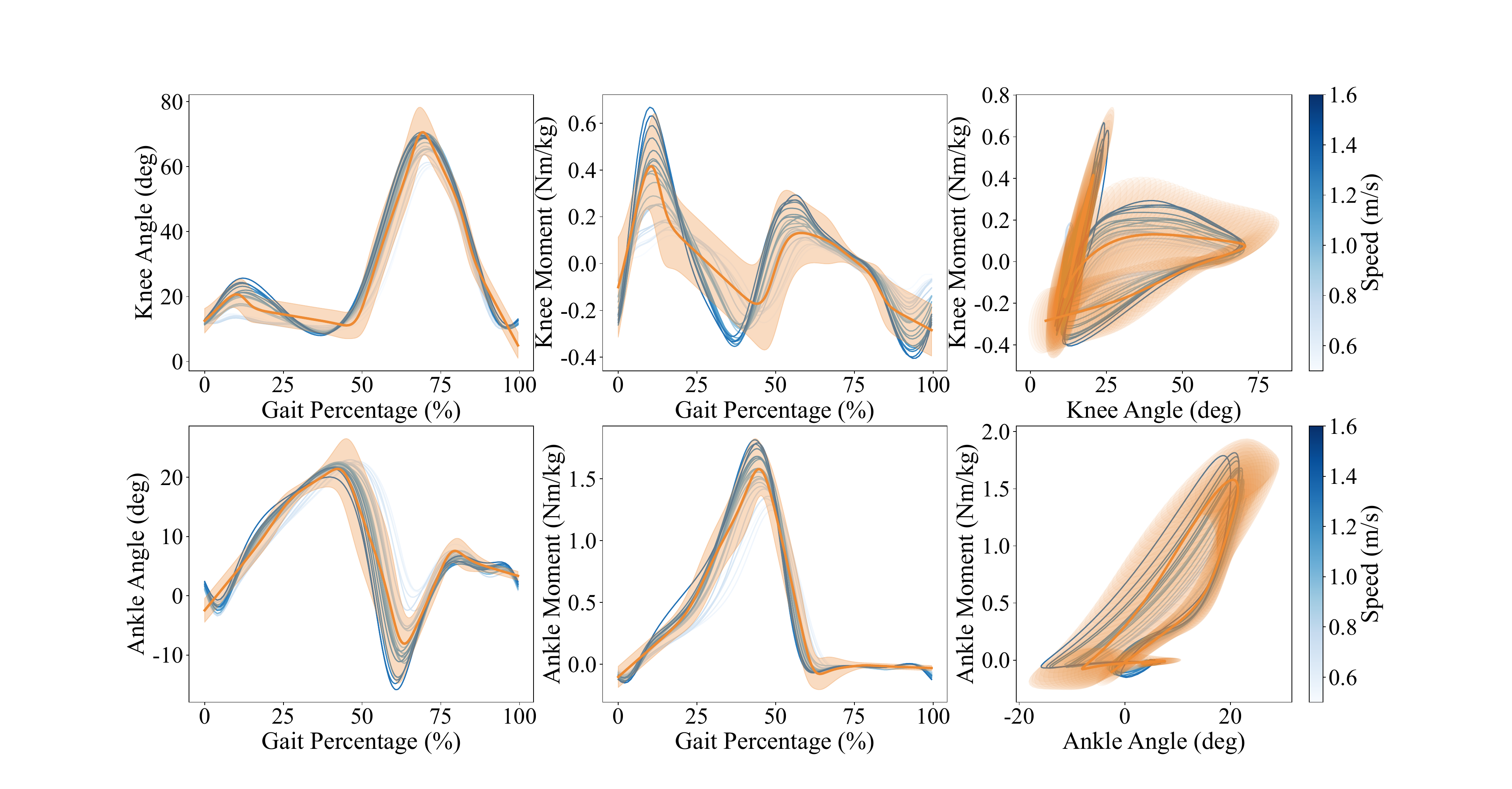}
	\centering
	\label{fig:5.1}}
	\subfigure[The torque-angle relationship with target features reconstructed by KMP.]{\includegraphics[width=0.49\linewidth]{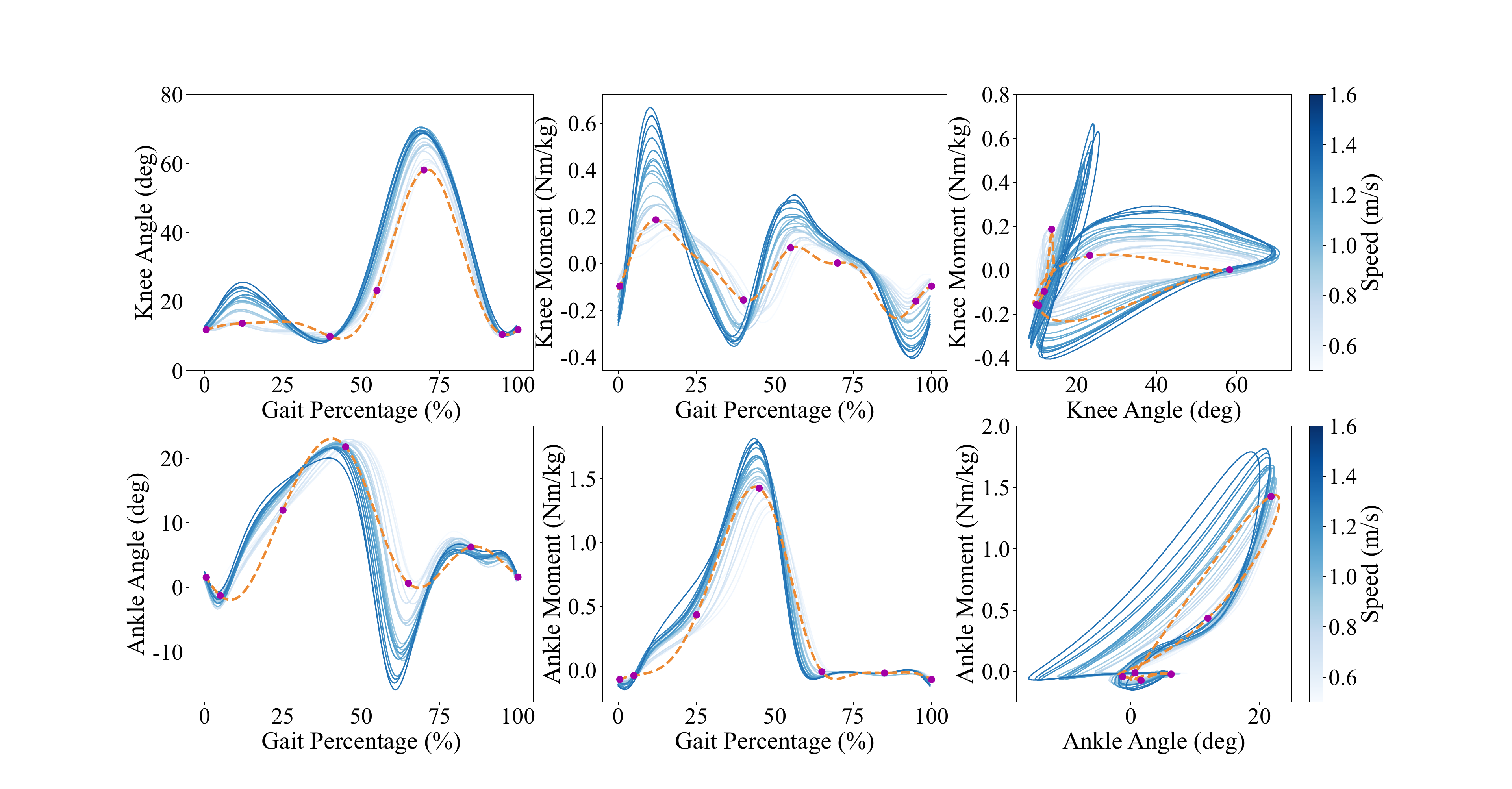}
	\centering
	\label{fig:5.2}}
	\caption{Experimental results of the KMP-based torque-angle relationship reconstruction. (a) Torque-angle relationship reproduces from human reference trajectories, where the blue curves denote the torque-angle relationships of human reference trajectories and the yellow curve (shadow regions denote variance) represents the reference torque-angle relationship retrieved by GMM-GMR. (b) Torque-angle relationship with target features is reconstructed by KMP, where the blue curves denote the torque-angle relationships of different walking speeds from human reference trajectories. The yellow dashed curve represents the torque-angle relationship passing through the target features and reconstructed by KMP at a new walking speed.} 
  \label{fig:5} 
\end{figure*} 

\begin{figure}[!t]
  \begin{center}
  \includegraphics[width=1.0\linewidth]{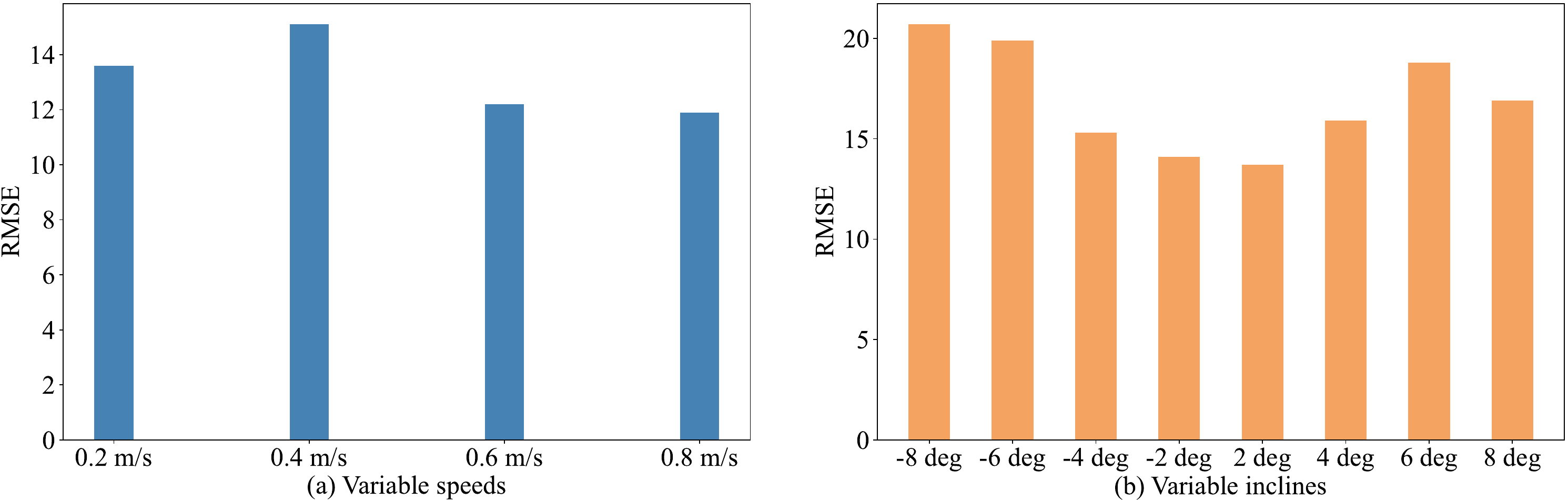}\\
  \caption{RMSE of the KMP model in the torque-angle relationship reconstruction.}
  \label{fig:6}
  \end{center}
\end{figure}

For the powered prostheses, the desired torque $\tau$ determined by the "quasi-stiffness" control law is defined as
\begin{equation}
\tau (\theta ) = \frac{{d{\tau _r}}}{{d{\theta _r}}}(\theta )(\theta  - {\theta _e})=k(\theta  - {\theta _e}),    
\end{equation}
where $k=\frac{{d{\tau _r}}}{{d{\theta _r}}}$ denotes the quasi-stiffness, $\theta_e$ represents the equilibrium angle.  For the prosthesis control, following the quasi-stiffness trends observed in able-bodied subjects \cite{lenzi2014speed}, we divide the gait phase into four sub-phases: stance flexion, stance extension, swing flexion, and swing extension. The sub-phase switching follows a finite-state machine (FSM), as shown in Fig. \ref{fig:3}. It is noted that the torque-angle relationship during the stance phase is approximately linear \cite{rouse2012difference,hansen2004human}. Inspired by this, as shown in Fig. \ref{fig:1}(d), this study applies a linear regression to estimate the quasi-stiffness of the torque-angle relationship, and the equilibrium angle is equal to the value of the intersection point between the regression equation and the x-axis. While the swing phase is much more complex than the stance phase \cite{hansen2004human}, we choose two linear parts that can achieve swing flexion and extension and use linear functions to approximate these two parts.

\begin{figure}[!t]
  \begin{center}
  \includegraphics[width=0.9\linewidth]{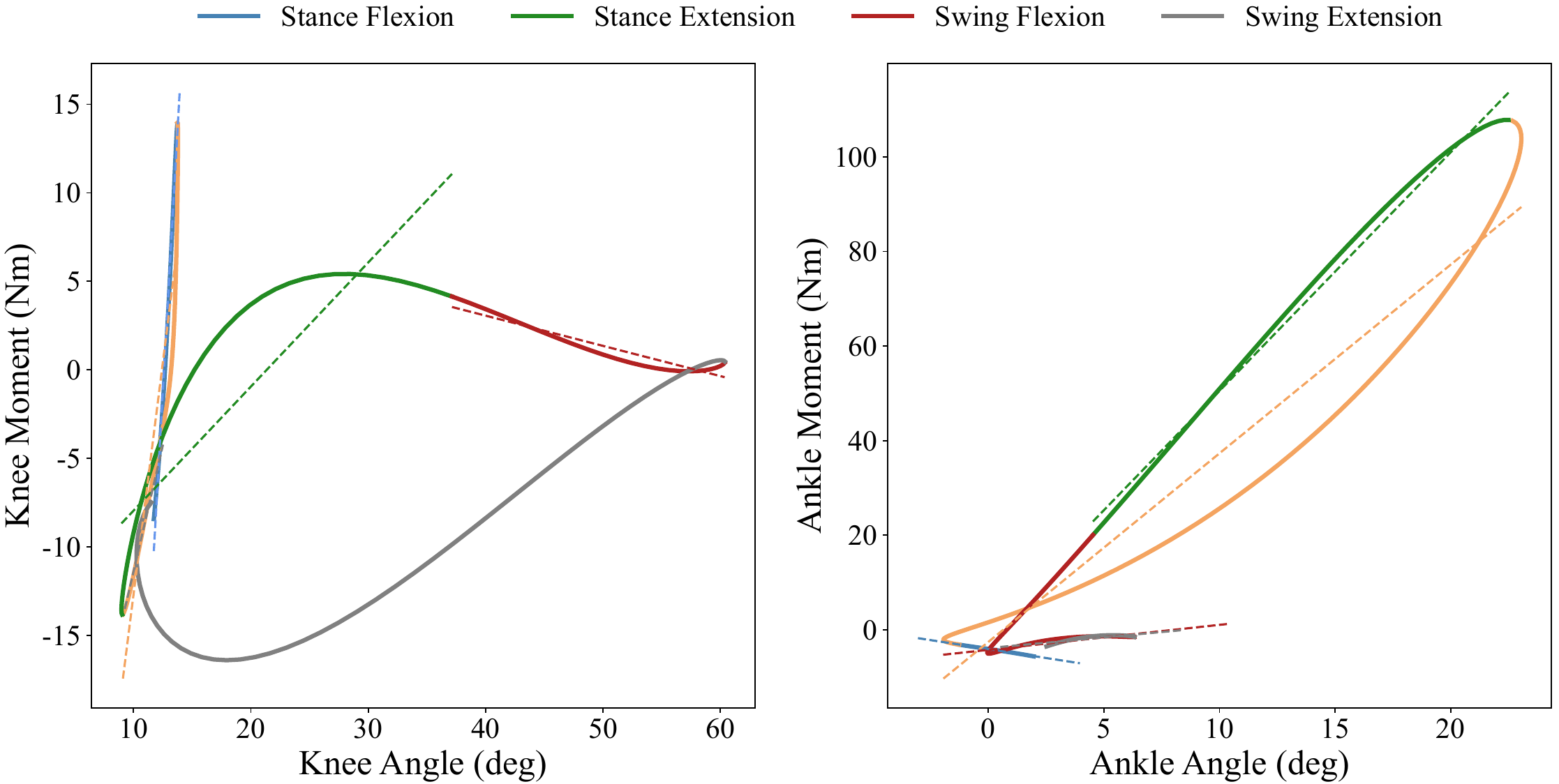}\\
  \caption{The quasi-stiffness estimation in 0.6 m/s walking task. The solid curves are the torque-angle relationship generated by the proposed KMP-based method, and each dash line denotes the linear regression of each sub-phase.}
  \label{fig:7}
  \end{center}
\end{figure}

\begin{figure}[!t]
  \begin{center}
  \includegraphics[width=1.0\linewidth]{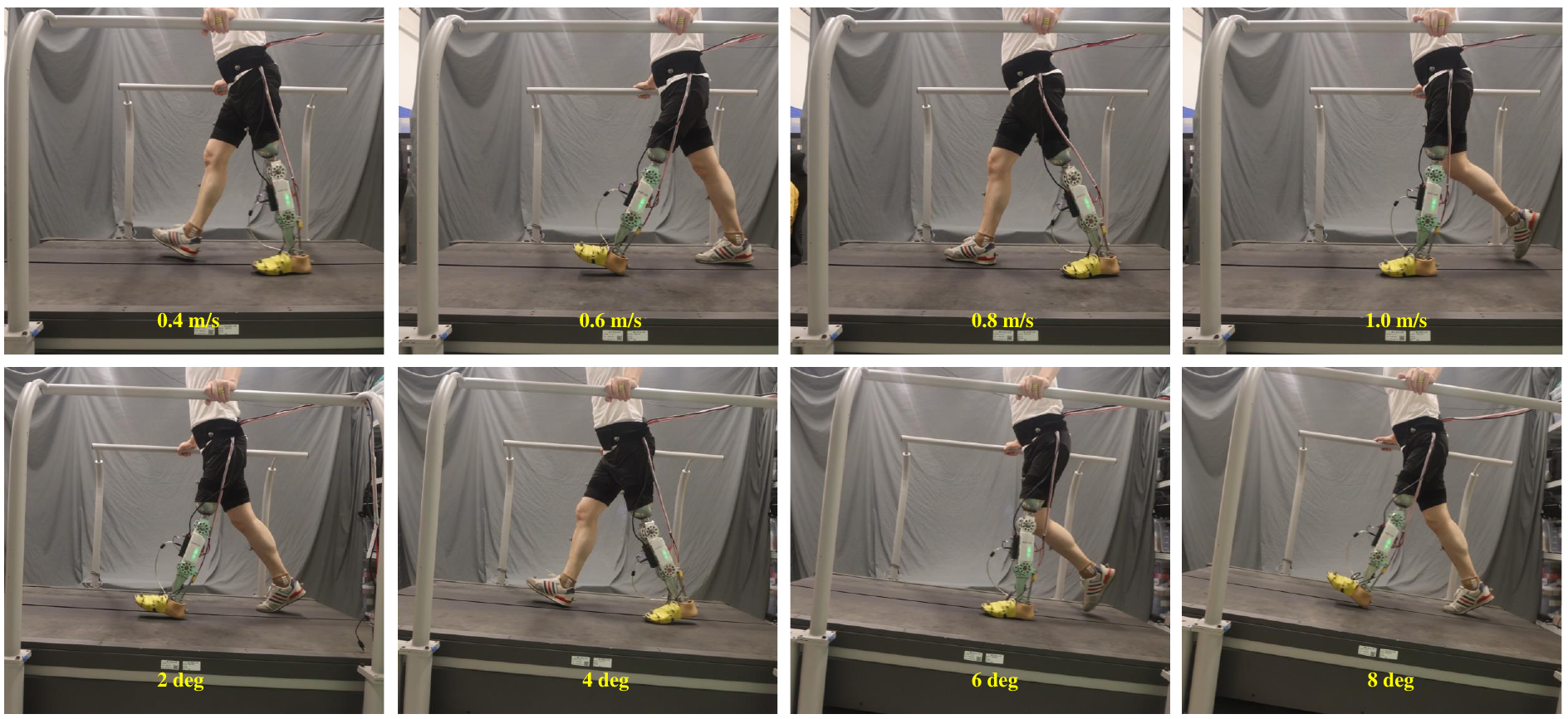}\\
  \caption{A series of gait trials of the participant performing various tasks.}
  \label{fig:8}
  \end{center}
\end{figure}

\begin{figure*}[!t]
  \begin{center}
  \includegraphics[width=1.0\linewidth]{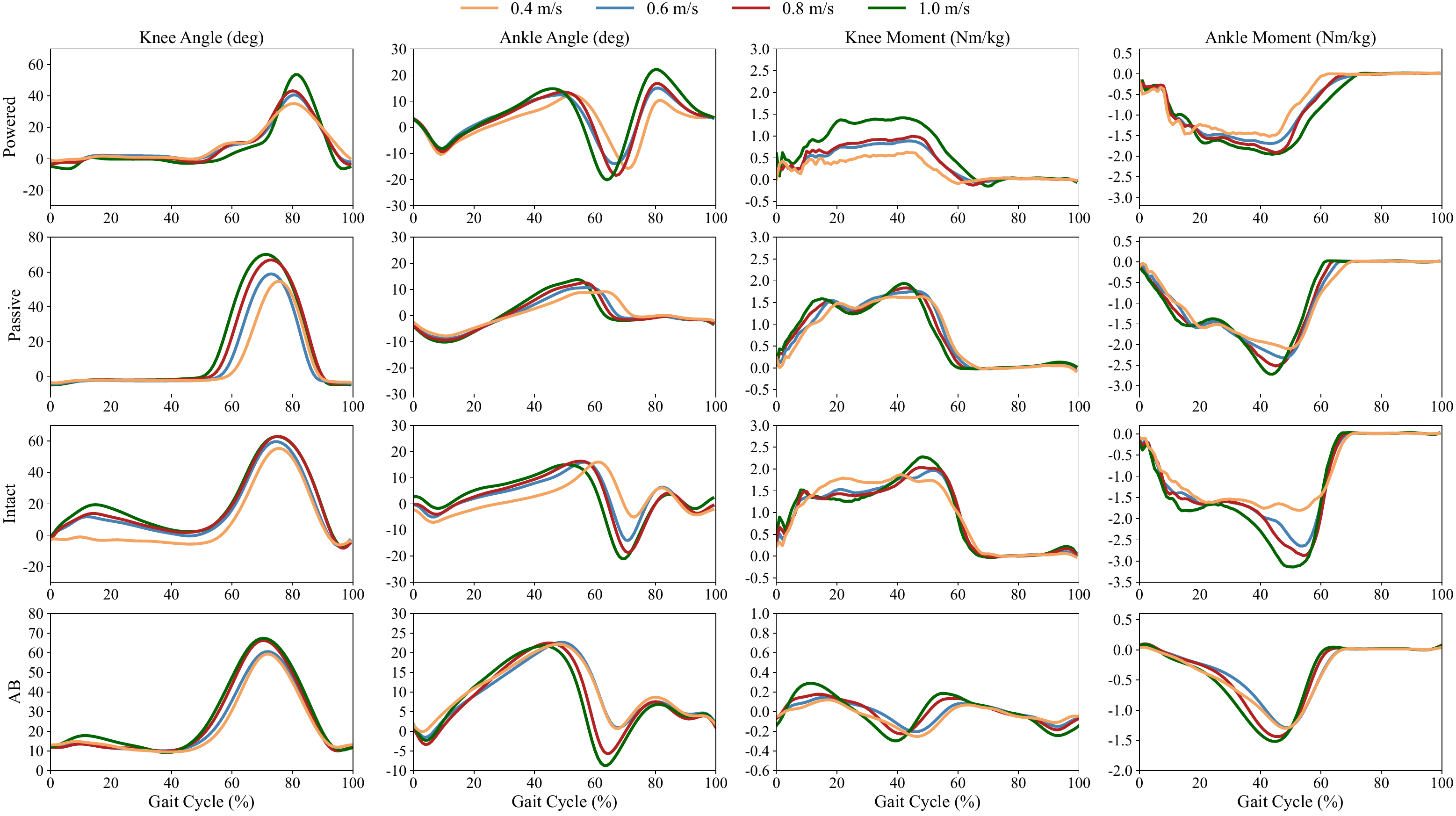}\\
  \caption{Mean kinematics and kinetics trajectories of the amputee participant performing varying speed task. (a) Observed joint kinematic and kinetic trajectories during the variable speed walking tasks. The first row shows the knee and ankle joint kinematics and kinetics generated by the powered transfemoral prosthesis. The second row shows the knee and ankle joint kinematics and kinetics of the passive prosthesis collected by the motion capture system. The third row shows the knee and ankle joint kinematics and kinetics of the intact leg while wearing the powered prosthesis under the motion capture system. The fourth row shows the able-bodied joint trajectories \cite{camargo2021comprehensive} in each task as a reference.}
  \label{fig:9}
  \end{center}
\end{figure*}

\begin{figure*}[!t]
	\centering
	\subfigure[Variable incline walking]{\includegraphics[width=1.0\linewidth]{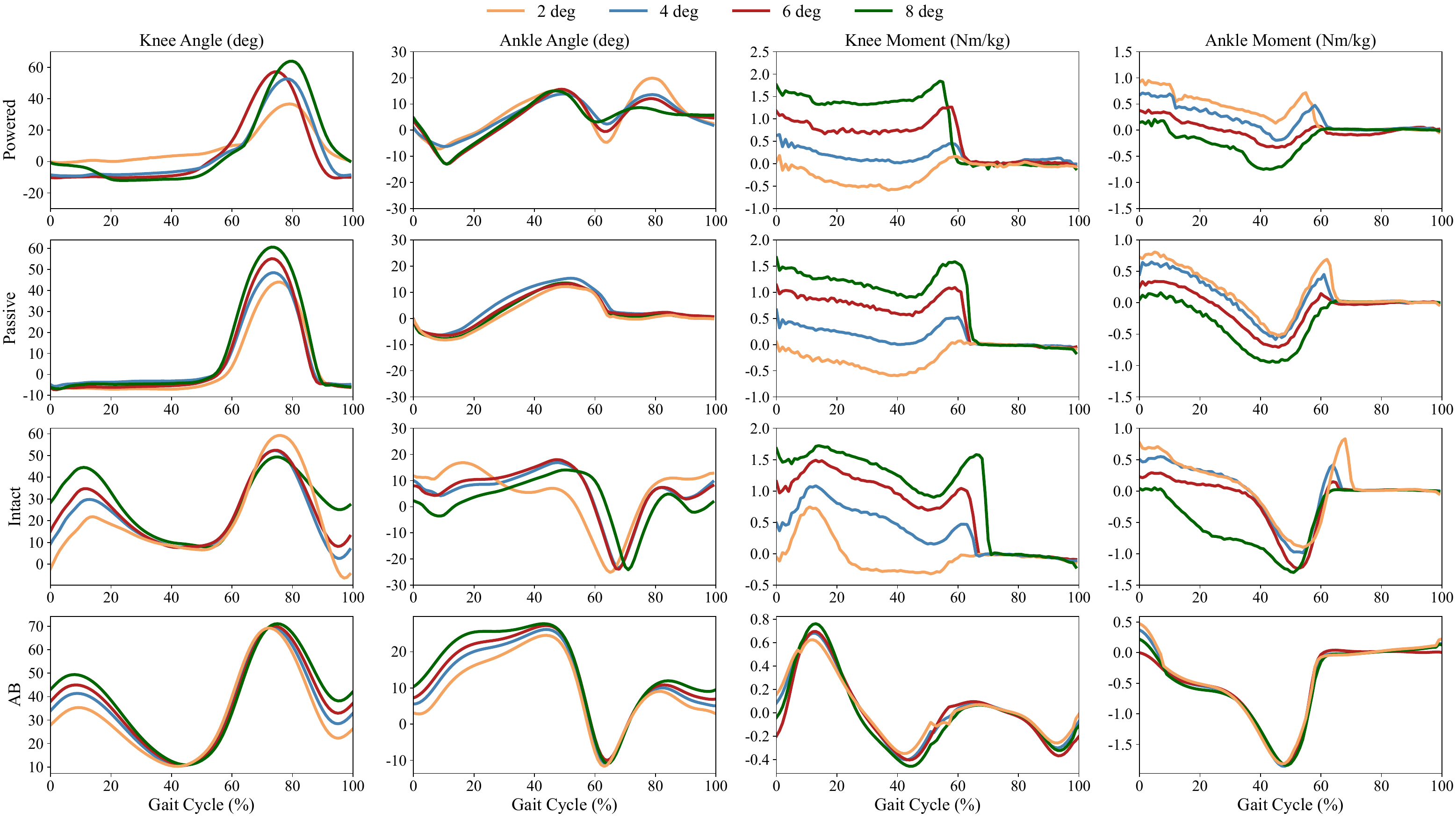}} 
	\centering
	\label{fig:10.1}
	\subfigure[Variable decline walking]{\includegraphics[width=1.0\linewidth]{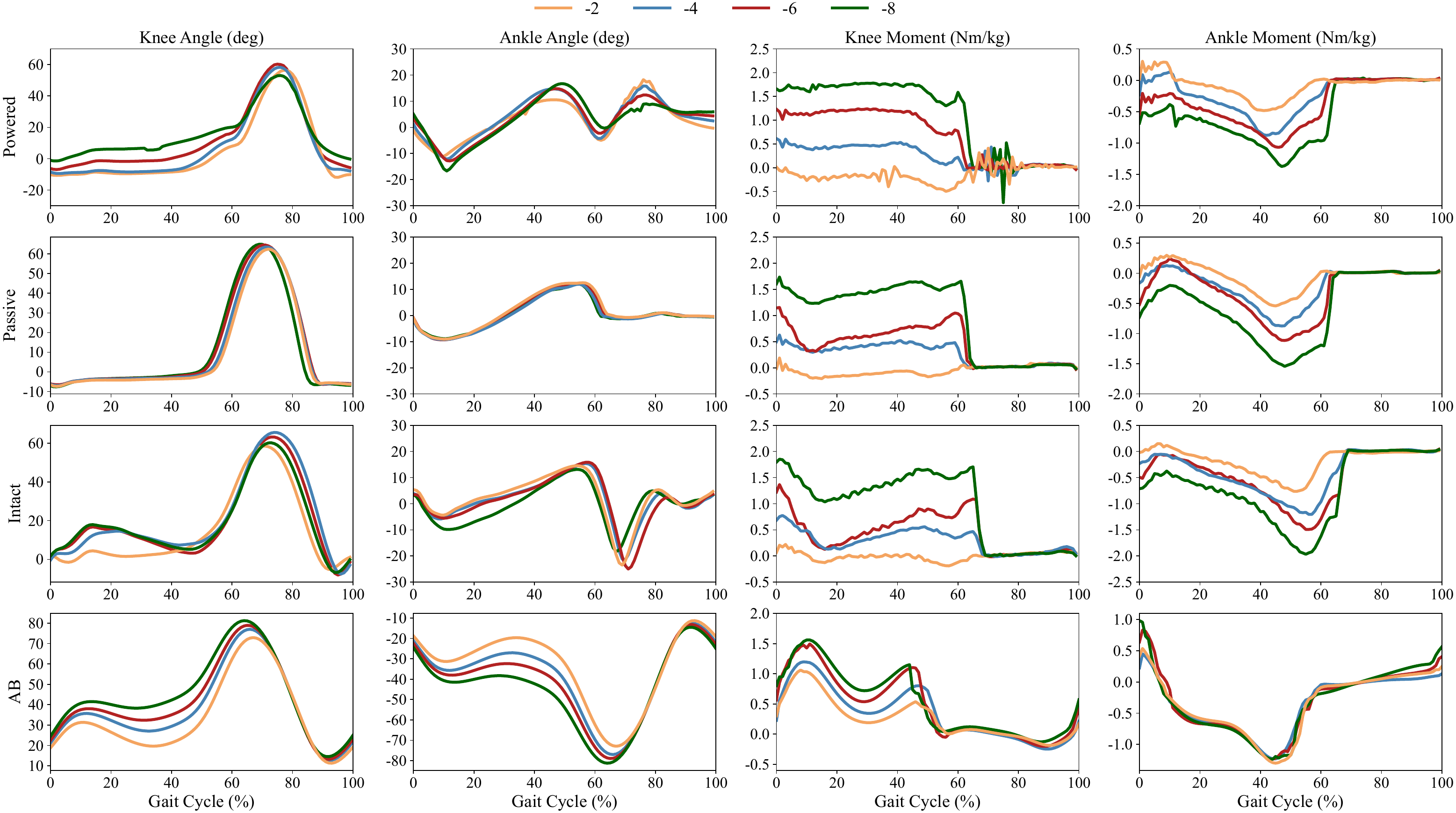}}
	\centering
	\label{fig:10.2}
	\caption{Mean kinematics and kinetics trajectories of the amputee participant performing varying varying incline tasks. (a) Observed joint kinematic and kinetic trajectories during the variable incline walking tasks. 
 (b) Observed joint kinematic and kinetic trajectories during the variable decline walking tasks.} 
  \label{fig:10} 
\end{figure*}

\begin{figure}[!t]
	\centering
	\subfigure[RMSE of kinematics]{\includegraphics[width=0.49\linewidth]{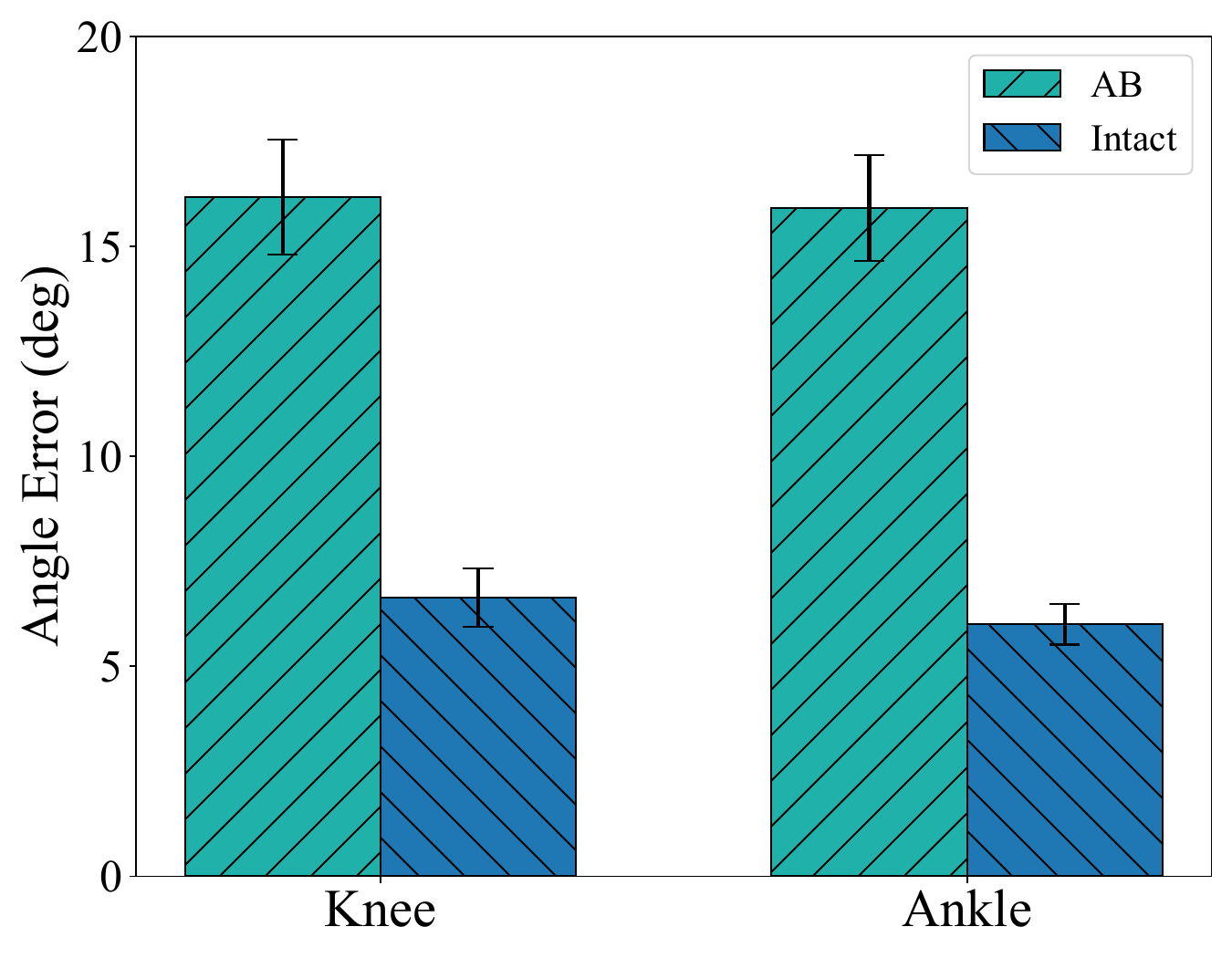}} 
	\centering
	\label{fig:11.1}
	\subfigure[RMSE of kinatics]{\includegraphics[width=0.49\linewidth]{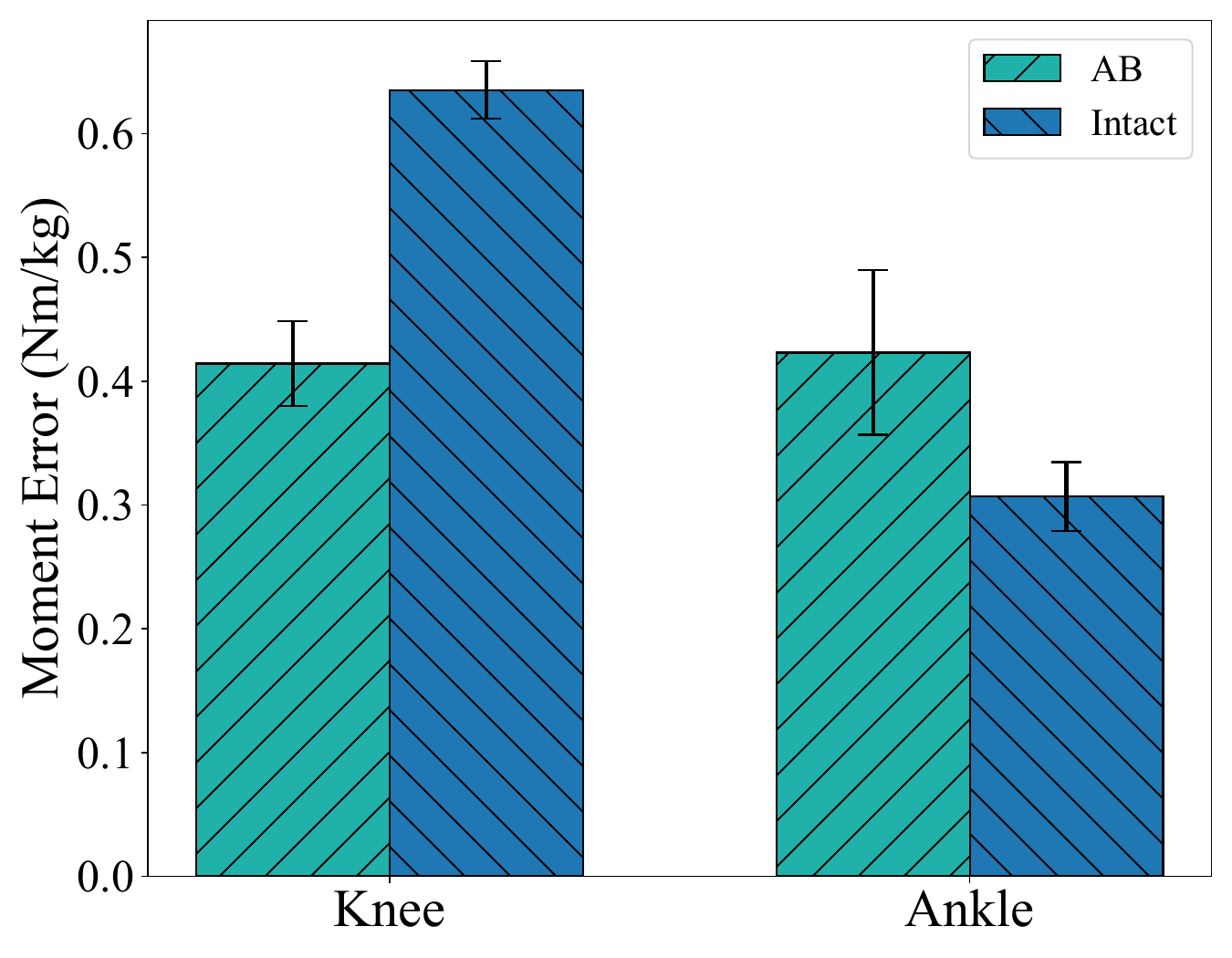}}
	\centering
	\label{fig:11.2}
	\caption{RMSE of the powered prosthetic joint kinematics (a) and kinetics (b) with respect to intact leg and able-bodied walking data during the experiments. The error bars denote a standard deviation.} 
  \label{fig:11} 
\end{figure} 

\begin{figure}[!t]
  \begin{center}
  \includegraphics[width=1.0\linewidth]{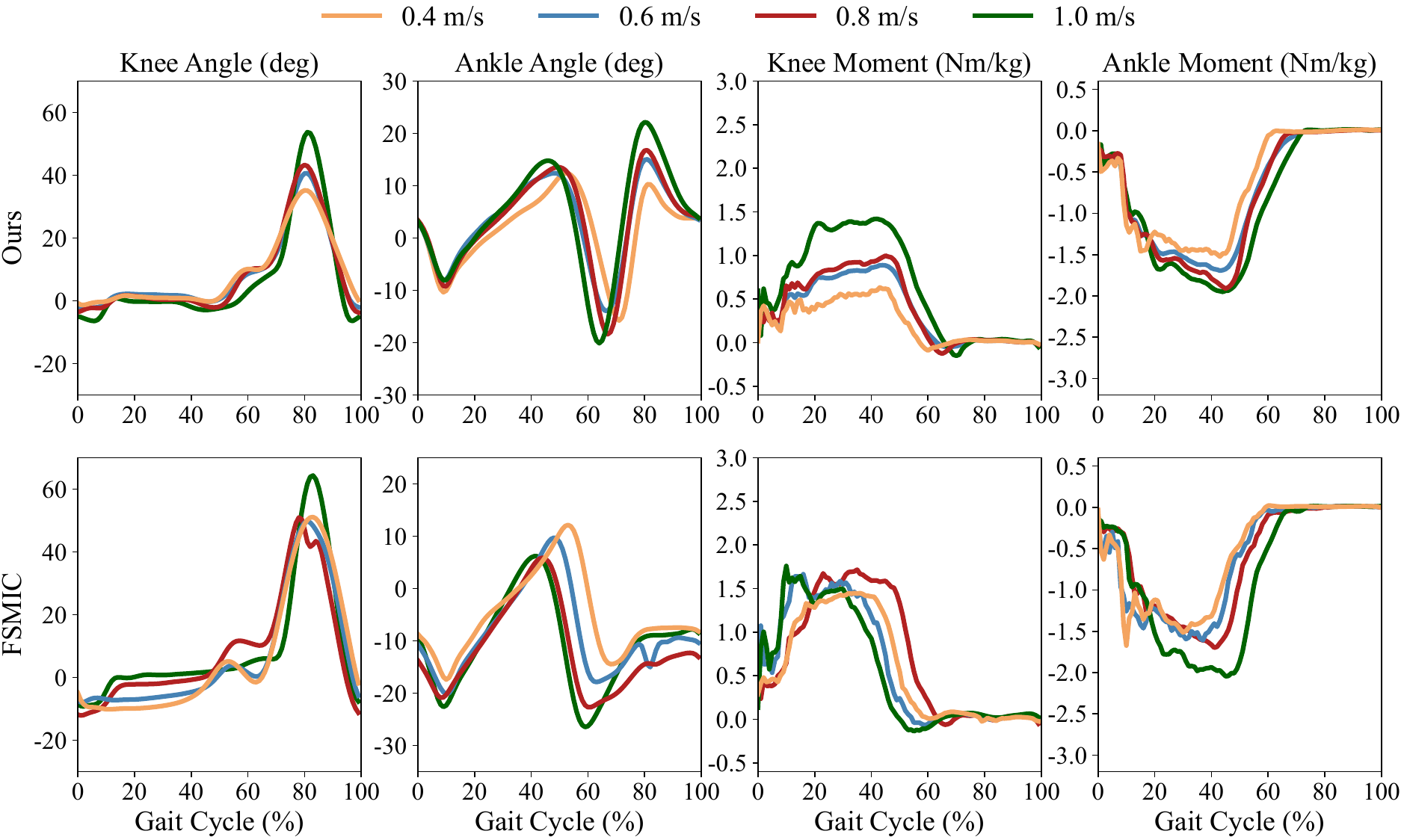}\\
  \caption{Mean kinematics and kinetics trajectories generated by the proposed learning quasi-stiffness controller and the benchmark FSMIC.}
  \label{fig:12}
  \end{center}
\end{figure}

\begin{figure}[!t]
	\centering
	\subfigure[RMSE of kinematics]{\includegraphics[width=0.49\linewidth]{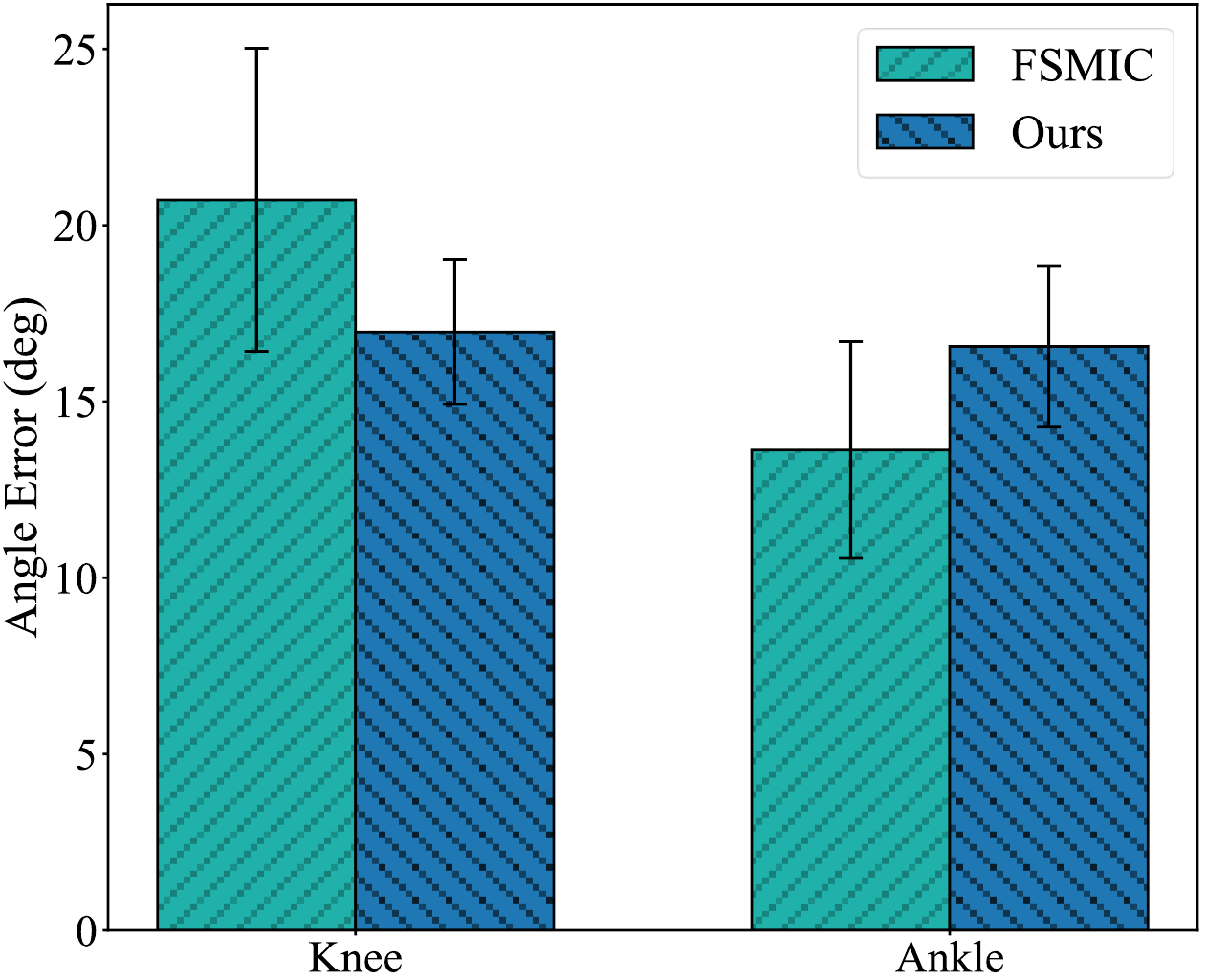}} 
	\centering
	\label{fig:13.1}
	\subfigure[RMSE of kinetics]{\includegraphics[width=0.49\linewidth]{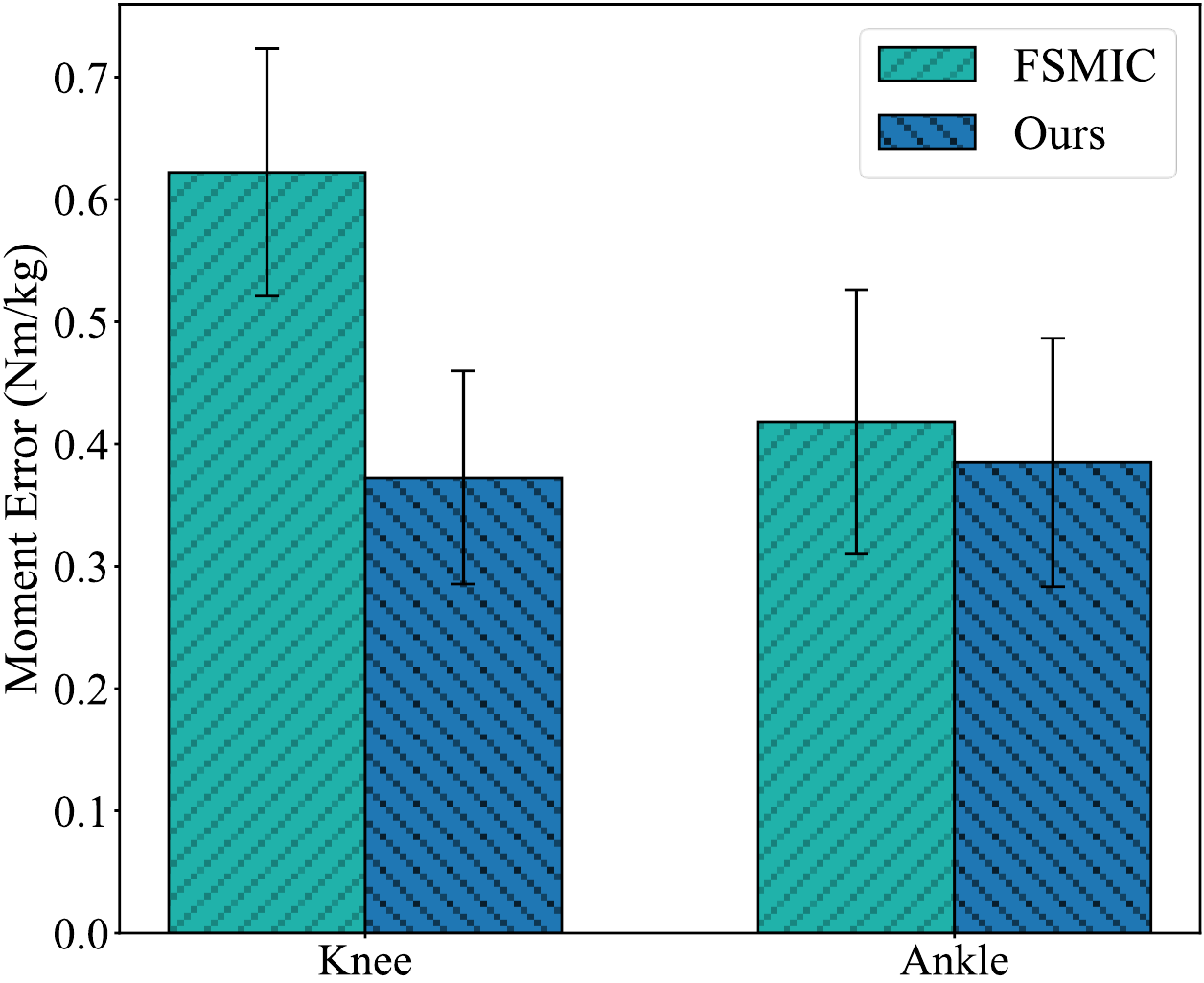}}
	\centering
	\label{fig:13.2}
	\caption{RMSE of the kinematics (a) and kinetics (b) generated by the proposed method and the benchmark FSMIC with respect to the able-bodied references during the experiments. The error bars denote a standard deviation.} 
  \label{fig:13} 
\end{figure} 

%After obtaining the torque-angle relationship of a new task, we investigated using linear functions to approximate several parts of the torque-angle relationship, such that the equations we chose can recapitulate the joint torques and angles of healthy people
% A linear regression function is introduced to estimate the quasi-stiffness and equilibrium angle for the quasi-stiffness control of the powered prostheses.

% % =======
% % FIG. 03
% % =======
% \begin{figure}[!t]
%   \begin{center}
%   \includegraphics[width=1.0\linewidth]{pdf/Figure_3_1.pdf}\\
%   \caption{Torque-angle relationship reproduces from human demonstrations, where the blue relationship denote the torque-angle relationship of human demonstrations and the yellow relationship (shadow regions denotes variance) represents the reference torque-angle relationship retrieved by GMM-GMR.}
%   \label{fig:4}
%   \end{center}
% \end{figure}

% % % =======
% % % FIG. 02
% % % =======
% \begin{figure}[!t]
%   \begin{center}
%   \includegraphics[width=1.0\linewidth]{pdf/Figure_3_new.pdf}\\
%   \caption{Torque-angle relationship with target features is reconstructed by KMP, where the blue relationship denotes the torque-angle relationship of different walking speeds from human demonstrations. The yellow dashed relationship represents a torque-angle relationship passing through the target features and reconstructed by KMP at a new walking speed.}
%   \label{fig:5}
%   \end{center}
% \end{figure}

% % =======
% % FIG. 03
% % =======

\section{Experiments and Results}

% % =======
% % FIG. 04
% % =======

% \begin{figure*}[!t]
%   \begin{center}
%   \includegraphics[width=1.0\linewidth]{pdf/lg.pdf}\\
%   \caption{Mean kinematics and kinetics trajectories of the transfemoral amputee performing varying speed tasks. Able-bodied trajectories \cite{camargo2021comprehensive} are also provided as a reference.}
%   \label{fig:7}
%   \end{center}
% \end{figure*}

% % =======
% % FIG. 05
% % =======

% \begin{figure*}[!t]
%   \begin{center}
%   \includegraphics[width=1.0\linewidth]{pdf/ra.pdf}\\
%   \caption{Mean kinematics and kinetics trajectories of the transfemoral amputee performing varying speed tasks. Able-bodied trajectories \cite{camargo2021comprehensive} are also provided as a reference.}
%   \label{fig:8}
%   \end{center}
% \end{figure*}

% \begin{figure}[!t]
%   \begin{center}
%   \includegraphics[width=1.0\linewidth]{pdf/Figure2.pdf}\\
%   \caption{ RMSE in the powered prosthetic joint kinematics (left) and kinetics (right) relative to able-bodied walking data during the steady-state task trials. The error bars represent a standard deviation over lumped participant strides.}
%   \vspace{-0.5cm}
%   \label{fig:8}
%   \end{center}
% \end{figure}

% \begin{figure*}[!t]
%   \begin{center}
%   \includegraphics[width=1.0\linewidth]{pdf/Figure1.pdf}\\
%   \caption{}
%   \label{fig:10}
%   \end{center}
% \end{figure*}
\subsection{Experimental Setup and Protocol}
1) \emph{Prosthesis Hardware}: The powered transfemoral prosthesis used in this study is shown in Fig. \ref{fig:3}. Both knee and ankle joints are driven by frameless motors (ILM $70 \times 16$ motor kit, RoboDrive, Seefeld, Germany) with a 50:1 harmonic reducer (CSD-25-50-2A-GR, Nagano Prefecture, Japan). The frameless motors are actuated by drives (G-MOLTWIR50/100SEO, Elmo Motion Control, Petah Tikva, Israel). A 6-axis force/torque sensor (Sunrise Instruments, Nanning, China) is employed to measure the GRFs (ground reaction forces). 

2) \emph{Experimental Protocol}: A transfemoral amputee patient (44 years old, 60 kg, 170 cm, 24 years post-amputation) is invited to participate in the experiments to evaluate the proposed learning quasi-stiffness control framework.
Notably, to evaluate the performance of the GPR model used in the target features estimation and the KMP model used in the torque-angle curve reconstruction, we used the walking speeds (from 0.2m/s to 0.8 m/s in 0.2 m/s increments) and ramp inclinations (from -8 deg to 8 deg in 2 deg increments) as test set. To investigate the task-adaptive performance of the proposed control framework, steady-state walking task trails including varying speeds (from 0.4 m/s to 1.0 m/s, $\pm 2$ m/s deviations) and varying inclines (from 2 deg to 8 deg, $\pm 2$ deg deviations) are conducted on an instrumented split-belt treadmill (Bertec, Columbus, OH, USA). The initial trials began with a period of acclimation in which the amputee participant first walked on the treadmill at a self-preferred speed (about 0.6 m/s) until feeling comfortable. After tuning and acclimation for a period of time, the participant walked on the treadmill again and performed each steady-state task trail for 30 seconds. The participant provided informed consent, and the Southern University of Science and Technology Institutional Review Board approved the experiments.

3) \emph{Benchmark FSMIC Setup}: For comparison, this article introduces a benchmark FSMIC based on the existing studies \cite{sup2008design,sup2010upslope}. The selection of this FSMIC as a benchmark stems from its extensive usage, simplicity, and capacity to reproduce human-like kinematics and kinetics after appropriate impedance tuning. During the benchmark FSMIC tuning period, combined with feedback from both the participants and the prosthetist, the tuning process persisted until the authors, prosthetist, and participant expressed satisfaction with the attained natural gait.
\subsection{Experimental Results}
Due to the space limitations, this study presents an example of varying walking speed tasks and varying inclines walking tasks to show how the learning quasi-stiffness control framework works. The learning process of the GPR-based target feature estimation and KMP-based torque-angle relationship construction is offline on the joint kinematics and kinetics obtained from the open-source dataset \cite{camargo2021comprehensive}.

1) \emph{Target Features Estimation}: Fig. \ref{fig:2}(b) shows the experimental results of target features prediction, where the blue dots denote the reference target features and the purple dots represent the predicted target features of a new task (new speed). The GPR-based target features prediction model is imported into the embedded platform of the powered prosthesis and used for feature prediction of new tasks. The prediction errors were calculated by
RMSE (root mean squared error) between reference and predicted target joint angles and joint torques, as shown in Fig.~\ref{fig:4}. 

2) \emph{Torque-angle Relationship Reconstruction}: 
Fig. \ref{fig:5} shows the torque-angle relationship learning and reconstruction in a new task. The distribution of the human reference torque-angle relationship can be learned by GMM. Based on the learned GMM, the reference torque-angle relationship can be reproduced by GMR. As shown in Fig. \ref{fig:5.1}, the blue torque-angle curves denote the human reference trajectoires of different speed walking tasks, and the yellow torque-angle curve represents the reference torque-angle relationship reproduced by GMR. As shown in Fig. \ref{fig:5.2}, the yellow dashed curve represents a torque-angle relationship reconstructed by the human reference torque-angle relationships (blue curves) and target features (purple dots) in a new task. The reconstruction errors were also calculated by
RMSE between reference and predicted torque-angle relationship, as shown in Fig.~\ref{fig:6}. 

3) \emph{Quasi-stiffness Estimation}: Fig. \ref{fig:7} shows using linear functions to approximate several parts of the knee and ankle torque-angle relationships in the varying speed walking tasks. 
%Table \ref{tab:1} shows the estimated stiffness and equilibrium angles from the torque-angle relationships of different speed tasks and these impedance parameters are used in the powered transfemoral prosthesis experiments.

4) \emph{Amputee Participant Experimental Results}: Fig. \ref{fig:8} shows a series of gait trails of a transfemoral amputee performing varying speed and inclines walking tasks. Fig. \ref{fig:9} and \ref{fig:10} present that the kinematic and kinetics trajectories produced by the proposed method are similar to the able-bodied references. The powered prosthetic joint trajectories show strong similarities to the intact joint trajectories, which also indicate that the quai-stiffness can improve gait symmetry. Fig. \ref{fig:11} shows the RMSE in the powered prosthetic joint trajectories relative to able-bodied trajectories and intact leg trajectories during the walking experiments. The lower RMSE of the observed kinematics with respect to the intact leg kinematics indicates that the proposed quasi-stiffness control achieves better joint kinematic symmetry. For comparison, Fig.~\ref{fig:12} shows the observed mean kinematics and kinetics trajectories of the proposed learning quasi-stiffness control framework and the benchmark FSMIC. Fig.~\ref{fig:13} shows the RMSE of the observed kinematics and kinetics generated by the proposed method and the benchmark FSMIC with respect to the able-bodied references. The angle and moment RMSE of the proposed method are usually smaller than the benchmark.

\section{Discussion}
\subsection{Evaluation of the Proposed Framework}
This work proposed a learning task-adaptive quasi-stiffness controller for a powered transfemoral prosthesis that generalizes across tasks. The proposed method has the potential to quickly and autonomously configure the impedance parameters for variable tasks. 
% Analysis of the robotic prosthesis kinematics (Figure 7) highlights the ability of the proposed control framework to fairly approximate the biomechanical behavior of intact legs at different walking speeds. 
Experimental results and walking performance (see the accompanying video) highlight its capacity to replicate able-bodied biomechanics across variable tasks. As shown in Fig. \ref{fig:9} and \ref{fig:10}, the observed kinematic and kinetic profiles of the powered prosthetic joint vary with walking tasks and are similar to able-bodied references, such as appropriately varying peak angles, peak moments, and trends of profiles. As speed or incline increases, the proposed method can generate biomechanical performances similar to the intact limb of the user and healthy individuals. For instance, as the incline increases, ankle joint torque correspondingly increases to provide individuals with limb amputations more push-off torque.
%(The knee and ankle profiles changed with tasks, as expected from the results of the stance-controller characterization)
The powered prosthetic joint trajectories show strong similarities to the intact joint trajectories (also the lower RMSE of the observed kinematics with respect to the intact leg kinematics shown in Fig. \ref{fig:11}(a)), which indicates that the quasi-stiffness control has the potential to improve gait symmetry. Knee kinematics show the most differences with respect to able-bodied data, which may be because the torque-angle relationship of the knee joint is more complex with fewer linear parts, which inevitably introduces errors when using linear approximations. 

This article introduces the widely used FSMIC controller as a benchmark for comparison. Fig. \ref{fig:12} and \ref{fig:13} show the proposed learning quasi-stiffness control framework meets the benchmark's performance in most walking tasks. Fig. \ref{fig:13}(b) indicates that the proposed method surpasses the benchmark's performance in generating the biological kinetics. As shown in Fig. \ref{fig:12} and Fig. \ref{fig:13}(a), the FSMIC’s performance was not drastically worse than the proposed method in the tested metrics, it required an average of 20 min of tuning per tuned task. The proposed method highlights its advantages in reproducing biomimetic kinematics and kinetics similar to able-bodied references, aligns with the performance of the FSMIC without requiring manual impedance tuning, and reveals the potential to enable autonomous task adaptation during continuously varying walking tasks.

\subsection{Limitations and Future Work}
% A limitation of this study lies in its absence of a direct comparison to previously presented powered prosthesis controllers. Nevertheless, the primary objective of this study was to propose and evaluate a task-adaptive quasi-stiffness control framework that generalizes across tasks. The validation of this control framework may serve as a basis for research into creating task-adaptive prosthesis controllers. Subsequent work will focus on expanding the control framework and evaluating the clinical advantages of the proposed approach in contrast to the state-of-the-art powered prosthesis controllers \cite{best2023data,cortino2023data,hood2022powered}. In addition, future studies will invite more participants to verify the ability of the proposed control framework to generalize across subjects. 

A limitation of this study is that the torque-angle relationship of the swing phase tends to be more nonlinear and complex as the task becomes more difficult (composite tasks, such as walking speed and incline changing simultaneously), which means that using linear regression to estimate quasi-stiffness may introduce more significant errors. In future work, we will investigate using nonlinear regression to estimate the swing phase quasi-stiffness. Nevertheless, this work focuses on designing a powered transfemoral prosthesis control framework that generalizes across various tasks. Experimental results and walking performance (see the accompanying video) preliminarily verify that the framework has the potential to achieve task-adaptive control. Future studies will focus on evaluating the performance of the proposed method in stair climbing tasks. Besides, we will invite more participants to verify the performance of the proposed framework and investigate the user preference of the proposed method and the FSMIC in different tasks.

\bibliographystyle{IEEEtran}
\bibliography{Bibliography}

\vfill


\end{document}